\pdfoutput=1

\documentclass[11pt]{article}

\usepackage[preprint]{coling}

\usepackage{times}
\usepackage{latexsym}

\usepackage[T1]{fontenc}

\usepackage[utf8]{inputenc}

\usepackage{microtype}

\usepackage{inconsolata}

\usepackage{graphicx}
\usepackage{amsmath}
\usepackage{multirow}
\usepackage{booktabs}
\usepackage{amssymb}
\usepackage{hyperref}

\title{RUIE: Retrieval-based Unified Information Extraction using Large Language Model}

\author{Xincheng Liao\textsuperscript{}, Junwen Duan\textsuperscript{}\thanks{\ \ Corresponding author. Email: \href{mailto:jwduan@csu.edu.cn}{jwduan@csu.edu.cn}}, Yixi Huang\textsuperscript{}, Jianxin Wang\textsuperscript{} \\
Hunan Provincial Key Lab on Bioinformatics, School of Computer Science and Engineering, \\
Central South University, Changsha, Hunan, China \\
\texttt{\{ostars, jwduan, yx.huang\}@csu.edu.cn, jxwang@mail.csu.edu.cn} \\
\href{https://github.com/OStars/RUIE}{https://github.com/OStars/RUIE}
}

\begin{document}
\maketitle
\begin{abstract}

Unified information extraction (UIE) aims to extract diverse structured information from unstructured text. While large language models (LLMs) have shown promise for UIE, they require significant computational resources and often struggle to generalize to unseen tasks. We propose RUIE (\textbf{R}etrieval-based \textbf{U}nified \textbf{I}nformation \textbf{E}xtraction), a framework that leverages in-context learning for efficient task generalization. RUIE introduces a novel demonstration selection mechanism combining LLM preferences with a keyword-enhanced reward model, and employs a bi-encoder retriever trained through contrastive learning and knowledge distillation. As the first trainable retrieval framework for UIE, RUIE serves as a universal plugin for various LLMs. Experimental results on eight held-out datasets demonstrate RUIE's effectiveness, with average F1-score improvements of 19.22 and 3.22 compared to instruction-tuning methods and other retrievers, respectively.

\end{abstract}

\section{Introduction}

\begin{figure}[ht]
    \centering
    \includegraphics[width=\linewidth]{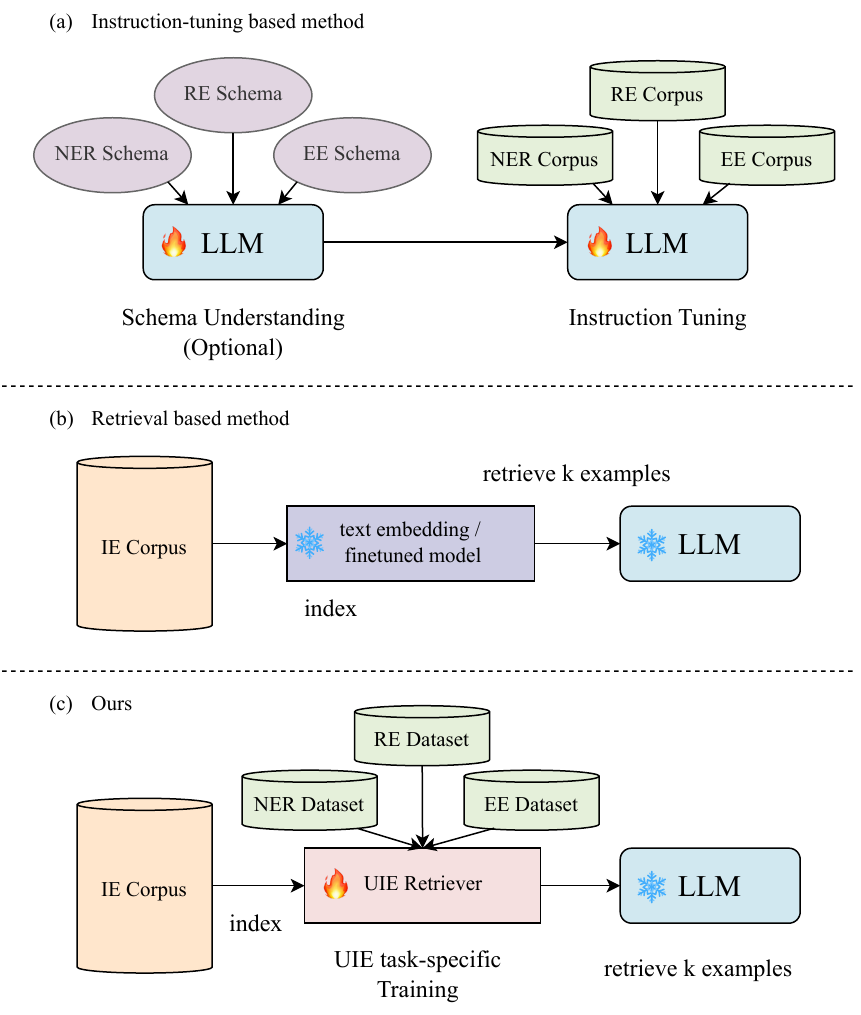}
    \caption{Illustration of three different paradigms for solving unified information extraction task.}
    \label{fig:methods_comp}
\end{figure}

Unified Information Extraction (UIE) represents a paradigm shift from traditional task-specific approaches, aiming to extract diverse structured information (e.g., Named Entity Recognition, Relationship Extraction, and Event Extraction) using a single model or framework. This unified approach demonstrates superior generalization capabilities and practical utility compared to conventional methods that require separate models for different extraction tasks \cite{lu_uie_2022, wang_instructuie_2023, li-etal-2024-knowcoder}. With the emergence of large language models (LLM) and their remarkable generalization abilities in various tasks \cite{cot, notellm, wang2024learningretrieveincontextexamples, medikal}, researchers have begun to explore LLM-based solutions to UIE challenges through two main approaches: instruction tuning and in-context learning (Figure \ref{fig:methods_comp}).

However, the inherent mismatch between structured information extraction outputs and LLM's pretraining format poses significant challenges, resulting in current large models still underperforming specialized IE approaches \cite{is_ie_solved_by_chatgpt, llm_reranker4hard, is_llm4ee}. Although some researchers \cite{wang_instructuie_2023, xiao_yayi-uie_2024, gui-etal-2024-iepile, sainz2024gollie, li-etal-2024-knowcoder} have bridged this gap by transforming IE annotations into textual or code pairs for instruction tuning, this approach faces several critical limitations: substantial computational costs, potential degradation of general capabilities \cite{xu-etal-2024-chatuie}, and limited generalization to unseen tasks.

Alternatively, in-context learning \cite{gpt3} has emerged as a promising direction, allowing LLMs to perform tasks with minimal examples or demonstrations. Recent works have demonstrated its effectiveness in IE tasks through various approaches: representing structured information in code format \cite{li_codeie_2023, wang_code4struct_2023}, improving extraction through offline sentence embedding and example retrieval \cite{guo_code4uie_2023}, and developing task-specific semantic representations \cite{wang_gpt-ner_2023, wan_gpt-re_2023}. Despite these advances, existing retrieval-based methods remain largely task-specific, lacking true UIE capabilities, and primarily relying on semantic relevance while overlooking LLMs' inherent preferences in example selection.


To address these limitations while leveraging LLMs' capabilities, we propose RUIE, a novel retrieval-based unified information extraction framework. As illustrated in Figure \ref{fig:methods_comp}, RUIE significantly reduces computational costs compared to instruction-tuning approaches by only requiring the fine-tuning of a smaller dense retriever (million-level parameters). Unlike existing retrieval-based methods, RUIE achieves true unified information extraction by maintaining a diverse candidate pool spanning multiple IE tasks (NER, RE, EE) and incorporating both semantic relevance and LLM preferences in example selection. Furthermore, we introduce a keyword-enhanced reward model to capture the label and fine-grained information, addressing the detailed nature of IE tasks \cite{wang_gpt-ner_2023, wan_gpt-re_2023, keyee, flr-mrc}. Our main contributions are as follows.

\begin{itemize}
\item We propose RUIE, a trainable retrieval framework for UIE that enables efficient task generalization through in-context learning while reducing computational costs.
\item We develop an innovative demonstration selection mechanism that uniquely combines LLM preferences with a keyword-enhanced reward model, enabling more accurate and context-aware example selection.
\item We demonstrate RUIE's strong flexibility as a general IE framework, showing robust performance across different tasks (NER, RE, EE) and easy integration with various LLM architectures.
\end{itemize}


\section{Related Works}

\subsection{Unified Information Extraction}
\cite{lu_uie_2022} first proposed the framework of unified Information Extraction. They used structured extraction language to unify the input and output forms of IE tasks and the structural schema instructor to guide the model generation, but it requires further fine-tuning for different downstream tasks. \cite{USM} considers the knowledge transfer between different tasks and schemas, modeling UIE as a semantic matching task with three dimensions of token-token, token-label and label-token, which achieves better zero/few-shot transfer ability. However, performing three-dimensional semantic matching for each token greatly increases the training and inference cost. Recently, some researchers have modeled IE as natural language \cite{wang_instructuie_2023, gui-etal-2024-iepile} or code \cite{sainz2024gollie, li-etal-2024-knowcoder} generation tasks. They constructed instruction-tuning datasets based on existing IE datasets to fine-tune large language models, realizing knowledge sharing of different tasks and effectively improving UIE performance. However, fine-tuning large language models is expensive, and the fine-tuned models do not generalize well in new domains.

\begin{figure*}[ht]
    \centering
    \includegraphics[width=\linewidth]{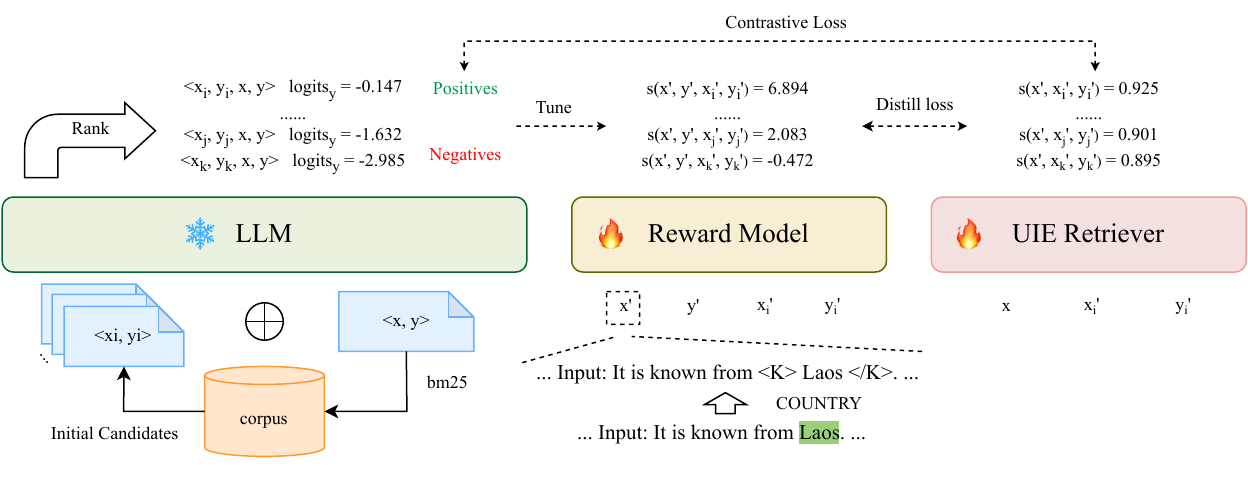}
    \caption{The overall architecture of RUIE. The training process consists of three steps: 1) the sparse retriever bm25 initializes a candidate set, which is then scored by the LLM. 2) a keyword-enhanced reward model captures fine-grained information. Keyword-enhanced strategy only applies to the input field of the example. 3) a bi-encoder dense retriever is trained using contrastive learning and knowledge distillation. During inference, the trained dense retriever selects the best demonstrations from the candidate pool $P$, and passes them to the LLM to produce the output. }
    \label{fig:method}
\end{figure*}

\subsection{In-context Learning based Information Extraction}
In-context learning (ICL) is the ability of large language models to perform new tasks with only a few examples or demonstrations. One significant merit of ICL is the circumvention of fine-tuning, which might not always be possible due to limited access to the model parameters or constraints on computational resources \cite{gpt3}. \cite{li_codeie_2023, wang_code4struct_2023} represent structured IE tasks with codes, and improve extraction performance by randomly selecting demonstrations. \cite{guo_code4uie_2023} optimizes the random retrieval process by introducing sentence embedding to select demonstrations according to the similarity between query and demonstrations. \cite{wang_gpt-ner_2023, wan_gpt-re_2023} show that fine-grained alignment information such as entities and relations is more important than sentence similarity for example selection in IE tasks. They use the entity and relation representations obtained by the fine-tuned small model to replace the sentence representation for retrieval, and obtain better performance than sentence-level embedding. However, they must obtain the entity span in the sentence and the fine-tuned small model in advance, which restricts the use cases. To the best of our knowledge, there is no trainable retrieval-based framework designed for UIE. On the one hand, we do not need any prior information about the input sentence, on the other hand, we design a retrieval training scheme for UIE, which achieves better performance than general sentence embedding.

\section{Methods}

In this section, we first present the formal definition of information extraction and the task setting for retrieval-based UIE. Then we present our training framework illustrated in Figure \ref{fig:method}.

\subsection{Problem Statement}
IE involves three main tasks: Named Entity Recognition (NER), Relation Extraction (RE), and Event Extraction (EE). For a given sentence $x$, NER seeks to extract tuples $\{s, e\}$, where $s$ represents the entity span and $e$ denotes the entity type. RE focuses on extracting triples $\{e_s, e_t, r\}$, with $r$ being the relation type and $e_s$ and $e_t$ indicating the head and tail entity, respectively. EE comprises two sub-tasks: Event Detection (ED) and Event Argument Extraction (EAE). ED involves extracting event triggers $t \in \mathcal{O}$, where $\mathcal{O}$ is the event type ontology, while EAE extracts arguments $a \in \mathcal{R}$ for a given event trigger $t$, with $\mathcal{R}$ being the role type ontology.

Given a target task (such as NER) test sample $x_{test}$ and k demonstrations ${(x_i, y_i)}_{i = 1}^k$, we use a frozen large language model to generate answer $y_{test}$ auto-regressively. Our goal is to retrieve $k$ demonstrations from the candidate pool $P$ that most closely match $y_{test}$ to $y_{truth}$. It is worth noting that our candidate pool $P$ contains a variety of information extraction tasks, such as NER, RE, ED and EAE.

\subsection{LLM Preference Scoring}

\textbf{Sample Format.} In the candidate pool $P$, each task sample $(x_i, y_i)$ consists of four parts: (1) Task Name: name of a specific IE task, such as ``Named Entity Recognition''. (2) Schema: task ontology presented in the form of python list. (3) Input: input context to be extracted. (4) Output: structured output linearized by natural language, such as ``Entitytype1: EntityName1; ...''. (see Appendix \ref{sec:sample_format}) 

\noindent\textbf{Score examples using LLM.} Previous retrieval-based information extraction methods usually only consider the text similarity between the input text and the candidate pool examples, ignoring the preference of the LLM. We assume that the LLM knows what good examples are; specifically, good examples can maximize the probability that the model produces ground-truth $y_{truth}$. Therefore, given an input $s = (x, y)$ in the training set, we enumerate each example $e_i = (x_i, y_i)$ in the candidate pool $P$. To keep with the inference stage, we concatenate the extraction instruction $I$ (see Appendix \ref{sec:instruction_format}), example $e_i$, and input $s$ into the model. We serve the token-level average log-likelihood of $y$ as the score between input $s$ and example $e_i$:
\begin{equation}
    Score(s, e_i) = \log~p(y | I; x_i; y_i; x)
\end{equation}

Then we rank the all candidate examples in descending order and select top $k$ and last $n$ as positive and negative examples respectively.

\noindent\textbf{Initialize candidates.} In practice, the candidate pool $P$ often comprises a significantly large number of examples. To reduce the costs of LLM scoring, given an input sample $s=(x, y)$, we use a sparse retriever bm25 \cite{bm25} to retrieve top-k candidate samples $\{(x_i, y_i)\}_i^k$ in the candidate pool $P$ as the set of scoring candidates. Choosing an appropriate k can greatly improve the scoring efficiency and ensure that the candidate set contains both positive and negative samples.

\subsection{Keyword-enhanced Reward}

\cite{wang_gpt-ner_2023, wan_gpt-re_2023} have shown that the alignment of query and candidate examples on fine-grained information such as entities and relations is more important than the alignment on coarse-grained information, like sentence semantics. Therefore, in order to make the fine-grained information between the input query and the candidate examples fully interactive, we propose a keyword-enhanced training strategy based on cross-encoder. On the one hand, the strategy aligns the fine-grained information through keyword, and on the other hand, the cross-encoder realizes the full interaction between query and candidate examples by accessing the ground-truth. Specifically, given a training sample $(x, y)$, for each information snippet $(sp, o)$ in $y$, where $o$ is the label of span $sp$, we add a pair of special tags ``\textless Keyword\textgreater'' and ``\textless/Keyword\textgreater'' around $sp$ in context $x$ (Figure \ref{fig:method}). Given an enhanced input $(x', y')$, we sample a positive example $(x'_{+}, y'_{+})$ in the top-k of the enhanced ranked candidates, take the last-n as the negative examples $(x'_{-_i}, y'_{-_i})$ and produce a real-valued score $s$. We train the cross-encoder using the cross-entropy loss:

\begin{small}
\begin{equation}
    \mathcal{L}_{reward} = -log\frac{e^{s(x',y',x'_{+}, y'_{+})}}{e^{s(x',y',x'_{+}, y'_{+})}+\sum_{i=1}^n{e^{s(x',y',x'_{-_i}, y'_{-_i})}}}
\end{equation}
\end{small}

\subsection{UIE Retriever Training}

In the test phase, we have no access to the ground-truth output $y$ corresponding to the input $x$. So we cannot directly use the above cross-encoder as the retriever. To enhance retrieval efficiency, we construct UIE retriever based on bi-encoder architecture. Specifically, we encode all samples $e'_i = (x'_i, y'_i)$ in the keyword-enhanced candidate pool $P$ into vector $h_{e'i}$ and build the index in advance. Given an input $x$, we use the same encoder to encode it into a vector $h_x$ and calculate matching score between the input $x$ and each example $e'_i$ using temperature-scaled dot product:
\begin{equation}
\begin{aligned}
    &score(x, e'_i) = h_x \cdot h_{e'i} / \tau \\
    h_{e'i} &= \text{avgpool}(Encoder(x'_i, y'_i)) \\
    h_x &= \text{avgpool}(Encoder(x))
\end{aligned}
\end{equation}

\noindent where ``avgpool'' refers to average pooling and $\tau$ refers to temperature hyperparameter.

We use two kinds of supervision signals to train the UIE retriever: (1) In order to make full use of the positive and negative pairs discovered by LLM, we use the Info-NCE loss $L_{contrastive}$ to perform contrastive learning between the top-k positives and in-batch negatives. (2) In order to make full use of the fine-grained alignment information of the reward model, we use the KL divergence to align the output distributions of the reward model and the retriever, calculated as $L_{distill}$, and the final training loss is the weighted sum of the above two losses:
\begin{equation}
    \mathcal{L}_{retriever} = \mathcal{L}_{distill} + \alpha\mathcal{L}_{contrastive}
\end{equation}

\noindent where $\alpha$ is a hyperparameter to measure the importance of the above two losses.

\begin{table*}[ht]
\centering
\resizebox{\textwidth}{!}{
\begin{tabular}{l|c|cc|c|ccc|c}
\toprule
\multirow{2}*{Method} & NER & \multicolumn{3}{c|}{RE} & \multicolumn{4}{c}{ED / EAE} \\
\cmidrule{2-9}
~ & CrossNER & FewRel & Wiki-ZSL & \#Avg & WikiEvents & RAMS & CrudeOil News & \#Avg    \\
\midrule
\multicolumn{9}{c}{Supervised Fine-tuning Methods}   \\
\midrule
UIE	&	38.37	&	-	&	-	&	-	&	5.12 / 1.78	&	9.25 / 2.14	&	6.45 / 8.95	&	6.94 / 4.29	\\
InstructUIE	&	49.36	&	39.55	&	35.20	&	37.38	&	11.64 / 5.88	&	24.27 / 6.21	&	23.26 / 21.78	&	19.72 / 11.29	\\
YAYI-UIE	&	50.39	&	36.09	&	41.07	&	38.58	&	10.97 / 5.11	&	18.87 / 8.21	&	12.45 / 19.74	&	14.10 / 11.02	\\
LLaMA2-IEPILE	&	56.50	&	37.14	&	36.18	&	36.66	&	13.93 / 12.55	&	23.62 / 11.30	&	33.87 / 18.47	&	23.81 / 14.11	\\
\midrule
\multicolumn{9}{c}{Retrieval-based Methods (k-shot=8)}   \\
\midrule
Random*	&	56.61	&	21.58	&	23.27	&	22.42	&	61.94 / 35.92	&	19.12 / 22.94	&	22.16 / 29.73	&	34.41 / 29.53	\\
BM25*	&	63.62	&	44.86	&	49.88	&	47.37	&	63.99 / 43.78	&	33.42 / 25.67	&	49.96 / 53.04	&	49.12 / 40.83	\\
\midrule
E5	&	45.21	&	41.50	&	44.62	&	43.06	&	66.09 / 41.18	&	26.78 / 25.49	&	46.47 / 53.12	&	46.45 / 39.93	\\
BGE	&	49.69	&	47.30	&	51.07	&	49.19	&	65.40 / \underline{43.12}	&	29.38 / 25.35	&	48.05 / 53.33	&	47.61 / \underline{40.60}	\\
BM25	&	59.57	&	44.27	&	50.72	&	47.50	&	63.79 / 40.11	&	25.46 / \underline{27.17}	&	\underline{54.05} / 53.48	&	47.77 / 40.25	\\
\midrule
RUIE	&	\underline{65.41}	&	\underline{49.93}	&	\underline{53.16}	&	\underline{51.55}	&	\underline{66.42} / 40.64	&	\underline{34.60} / 26.06	&	51.50 / \underline{53.79}	&	\underline{50.84} / 40.16	\\
RUIE-Deepseek	&	\textbf{69.60}	&	\textbf{57.57}	&	\textbf{60.14}	&	\textbf{58.85}	&	\textbf{71.88} / \textbf{44.30}	&	\textbf{47.72} / \textbf{38.51}	&	\textbf{68.73} / \textbf{61.04}	&	\textbf{62.77} / \textbf{47.95}	\\
\bottomrule
\end{tabular}
}
\caption{Performance (in F1-score) on NER, RE, ED and EAE tasks under held-out settings. Bold indicates the highest scores and the second-best scores are underlined. ``BM25*'' and ``Random*'' refer to using BM25 or randomly selecting demonstrations from the candidate pool corresponding to the task, rather than from the multi-task candidate pool. The only difference between RUIE and other retrieval-based methods is that they use different retrievers. It's worth noting that the model size of InstuctUIE and LLaMA2-IEPILE are 11B and 13B, respectively, while the default model size of RUIE is 8B.}
\label{tab:heldout_results}
\end{table*}

\section{Experiment Setup}

\subsection{Datasets}

To exhaustively evaluate the generalization ability of RUIE, we collect 31 held-in and 8 held-out datasets. We used the training set of the held-in and held-out datasets to form the candidate pool $P$. We constructed the retriever training set based on the held-in dataset, specifically, we sampled 10000 samples from each dataset and included all examples from datasets with less than 10000 samples. The Held-out dataset samples was completely unseen during training, and the UIE Retriever was tested on the test set of held-in and held-out datasets after training.

\subsection{Metrics}
We employ span-based Micro-F1 to evaluate the performance of our method. For NER, an entity is considered correct if the entity span and type are correctly predicted. For RE, a relation is considered correct if relation type, subject entity, and object entity match the golden annotation. For EE task, we report two evaluation metrics: (1) ED: an event trigger is correct if the event type and the trigger word are correctly predicted. (2) EAE: an event argument is correct if its role type and event type match a reference argument mention.


\subsection{Baseline Methods}
We compare our approach with three categories of methods:

\textbf{Small-PLM based methods}: \cite{lu_uie_2022} proposed a unified framework based on medium-sized language models with task-specific instructions and structured prediction.

\textbf{Instruction-tuning methods}: These approaches fine-tune large language models with task-specific instructions. \cite{wang_instructuie_2023} reformulated the IE tasks using natural language instructions. \cite{gui-etal-2024-iepile} and \cite{xiao_yayi-uie_2024} developed large-scale instruction datasets with JSON-formatted outputs. \cite{li-etal-2024-knowcoder} encoded IE structures through code-like formats.

\textbf{Retrieval-based methods}: We include both sparse and dense retrievers. For sparse retrieval, we use BM25~\cite{bm25}. For dense retrieval, we employ state-of-the-art embedding models E5~\cite{e5} and BGE~\cite{bge}, which are trained through contrastive learning on large-scale text pairs.
We exclude methods that require pre-identified entity spans~\cite{wang_gpt-ner_2023, wan_gpt-re_2023} to ensure fair comparison under our end-to-end extraction setting.


\subsection{Implementation Details}
We employ LLaMA3-8B as the scoring model and LLaMA3.1-8B-instruct for inference in our experiments by default. For each test instance, we retrieve 8 demonstrations by default. The keyword-enhanced reward model and retriever are implemented based on ELECTRA-base and E5-base, respectively. Detailed hyperparameters and training configurations are provided in Appendix \ref{sec:implementation_details}.

\section{Results and Analyses}

\subsection{Generalization to Held-out Tasks}
We evaluate the generalization capability of different approaches on held-out tasks. The SFT-based methods are directly tested on held-out tasks after training on held-in tasks. RUIE first trains on held-in tasks, then retrieves demonstrations from the mixed candidate pool for LLM on held-out tasks. The remaining retrieval-based methods directly retrieve demonstrations from the mixed candidate pool for LLM on held-out tasks. The results are presented in Table \ref{tab:heldout_results}.

\textbf{RUIE achieves the best performance in four information extraction tasks, and demonstrates rapid generalization to new tasks.} Compared with SFT-based methods, the generalization ability of RUIE is significantly better. The performance of SFT-based methods decreases seriously on datasets not seen during training, with the drop becoming more pronounced as task difficulty increases. For example, in the ET and EAE tasks, the best performing LLaMA2-IEPILE model only achieves F1 scores of 23.81 and 14.11, making it impractical for use in the real world. In contrast, RUIE delivers improvements of 8.91, 14.89, 27.03, and 26.05 on NER, RE, ET, and EAE tasks, respectively, despite using a smaller model.

\textbf{RUIE retrieves higher-quality examples compared to general-purpose retrievers.} BM25 is a strong baseline and outperforms semantic similarity-based retrieval methods on average, confirming that fine-grained information alignment is more important in IE example retrieval. By explicitly modeling fine-grained information, RUIE introduces LLMs preference as a supervision signal during retrieval training. Compared to BM25, which has the best overall performance, RUIE achieves 5.84, 4.05, and 3.07 improvements on the NER, RE, and EET tasks, respectively.

\textbf{RUIE demonstrates superior performance compared to simpler single-task corpus setups.} When using general retrievers to retrieve examples for NER, there is a risk of retrieving RE or EE examples. In contrast, RUIE effectively minimizes the likelihood of retrieving examples from unrelated tasks. We conducted an experiment using BM25 to search within the candidate pool of the same task. Despite this simpler setup, RUIE achieved improvements of 1.79, 4.18, and 1.72 on the NER, RE, and EET tasks, respectively, even in multi-task candidate pool.

However, RUIE does not achieve the performance improvement on EAE. We believe there are two main reasons. First, compared to other IE tasks, event extraction data is more scarce, leading to insufficient training during the pre-training and alignment stages. Additionally, EAE requires the model to extract event parameters while understanding the event structure, making it more challenging than other tasks. Furthermore, the amount of EAE training data is less than other subtasks, limiting RUIE's generalization performance on new EAE tasks.

\begin{table}[h]
\centering
\setlength{\tabcolsep}{2pt}
\begin{tabular}{l|c|c|c|c}
\toprule
Dataset & SOTA & E5 & BM25 & RUIE    \\
\midrule
ACE 2004	&	$\blacktriangledown$ 87.60	&	42.79	&	48.46	&	56.53	\\
ACE 2005	&	$\bigstar$ 86.66	&	42.53	&	48.35	&	55.86	\\
AnatEM	&	$\bigstar$ 90.89	&	41.71	&	46.85	&	51.58	\\
bc2gm	&	$\bigstar$ 85.16	&	43.48	&	45.99	&	49.78	\\
bc4chemd	&	$\spadesuit$ 90.56	&	48.29	&	50.39	&	55.03	\\
bc5cdr	&	$\bigstar$ 89.59	&	71.41	&	72.76	&	74.49	\\
Broad Tweet	&	$\blacktriangle$ 83.52	&	61.92	&	56.83	&	69.25	\\
CoNLL 2003	&	$\blacktriangle$ 96.77	&	68.09	&	68.50	&	78.34	\\
FabNER	&	$\clubsuit$ 82.90	&	35.24	&	39.61	&	38.51	\\
FindVehicle	&	$\spadesuit$ 99.45	&	68.86	&	73.53	&	92.26	\\
GENIA\_NER	&	$\spadesuit$ 78.29	&	53.23	&	57.64	&	59.85	\\
HarveyNER	&	$\bigstar$ 88.79	&	29.74	&	33.59	&	37.72	\\
MultiNERD	&	$\clubsuit$ 96.10	&	82.50	&	81.71	&	88.97	\\
ncbi	&	$\bigstar$ 90.23	&	54.60	&	59.14	&	58.81	\\
Ontonotes	&	$\bigstar$ 90.19	&	48.31	&	49.85	&	62.94	\\
PolyglotNER	&	$\blacktriangle$ 70.85	&	49.15	&	49.21	&	53.43	\\
TweetNER7	&	$\blacktriangle$ 66.99	&	49.93	&	49.79	&	53.17	\\
WikiANN en	&	$\clubsuit$ 87.00	&	64.07	&	58.27	&	68.81	\\
WikiNeural	&	$\bigstar$ 91.36	&	74.58	&	73.28	&	81.35	\\
\midrule
Avg	&	86.99	&	54.23	&	55.99	&	62.46	\\
\bottomrule
\end{tabular}
\caption{Performance (in F1-score) comparison on NER tasks under held-in setting. $\blacktriangledown$ indicates \cite{lu_uie_2022}, $\bigstar$ indicates \cite{wang_instructuie_2023}, $\spadesuit$ indicates \cite{gui-etal-2024-iepile}, $\blacktriangle$ incicates \cite{xiao_yayi-uie_2024}, $\clubsuit$ indicates \cite{li-etal-2024-knowcoder}.}
\label{tab:supervised_ner}
\end{table}

\begin{table}[h]
\centering
\setlength{\tabcolsep}{3.5pt}
\begin{tabular}{l|c|c|c|c}
\toprule
Dataset & SOTA & E5 & BM25 & RUIE    \\
\midrule
ADE corpus	&	$\spadesuit$ 82.31	&	70.39	&	65.44	&	71.24	\\
Conll04	&	$\bigstar$ 78.48	&	41.42	&	44.52	&	54.61	\\
GIDS	&	$\bigstar$ 81.98	&	27.49	&	30.44	&	40.03	\\
Kbp37	&	$\clubsuit$ 78.00	&	12.76	&	11.79	&	19.78	\\
NYT	&	$\spadesuit$ 94.04	&	55.95	&	74.85	&	72.30	\\
NYT11	&	$\blacktriangle$ 57.53	&	31.16	&	33.49	&	40.92	\\
SciERC	&	$\spadesuit$ 45.89	&	12.25	&	16.62	&	20.36	\\
Semeval RE	&	$\bigstar$  73.23	&	23.07	&	21.84	&	36.77	\\
\midrule
Avg	&	73.93	&	34.31	&	37.37	&	44.50	\\
\bottomrule
\end{tabular}
\caption{Performance (in F1-score) comparison on RE tasks under held-in setting 
$\bigstar$ indicates \cite{wang_instructuie_2023}, $\spadesuit$ indicates \cite{gui-etal-2024-iepile}, $\blacktriangle$ incicates \cite{xiao_yayi-uie_2024}, $\clubsuit$ indicates \cite{li-etal-2024-knowcoder}.}
\label{tab:supervised_re}
\end{table}

\begin{table}[h]
\centering
\setlength{\tabcolsep}{4.5pt}
\begin{tabular}{c|c|c|c|c}
\toprule
Dataset & SOTA & E5 & BM25 & RUIE    \\
\midrule
ACE2005	&	$\bigstar$ 77.13	&	42.90	&	39.06	&	53.41	\\
CASIE	&	$\bigstar$ 67.80	&	29.35	&	39.69	&	40.46	\\
PHEE	&	$\bigstar$ 70.14	&	37.18	&	59.21	&	47.16	\\
\midrule
Avg	&	71.69	&	36.48	&	45.99	&	47.01	\\
\bottomrule
\end{tabular}
\caption{Performance (in F1-score) comparison on ED tasks under held-in setting. $\bigstar$ indicates \cite{wang_instructuie_2023}.}
\label{tab:supervised_et}
\end{table}

\begin{table}[h]
\centering
\setlength{\tabcolsep}{4.5pt}
\begin{tabular}{c|c|c|c|c}
\toprule
Dataset & SOTA & E5 & BM25 & RUIE    \\
\midrule
ACE2005	&	$\bigstar$ 72.94	&	39.06	&	38.29	&	44.29	\\
CASIE	&	$\blacktriangle$ 64.23	&	39.69	&	40.85	&	43.76	\\
PHEE	&	$\blacktriangle$ 77.19	&	59.21	&	59.10	&	65.13	\\
\midrule
Avg	&	71.45	&	45.99	&	46.08	&	51.06	\\
\bottomrule
\end{tabular}
\caption{Performance (in F1-score) comparison on EAE tasks under held-in setting. $\bigstar$ indicates \cite{wang_instructuie_2023}, $\blacktriangle$ incicates \cite{xiao_yayi-uie_2024}.}
\label{tab:supervised_eea}
\end{table}

\subsection{Results on Held-in Tasks}
In the held-in tasks experiment, both the supervised fine-tuning methods and RUIE are trained on held-in tasks and tested on their corresponding test sets. The remaining retrieval-based methods directly retrieve k examples from held-in tasks for LLM. SFT-based methods include \cite{lu_uie_2022, wang_instructuie_2023, gui-etal-2024-iepile, xiao_yayi-uie_2024, li-etal-2024-knowcoder}. As some baselines only report results for specific IE tasks, we report the SOTA results of the above methods in each dataset, denoted as ``SOTA'' in the tables.

The results of NER, RE, ED and EAE are shown in Tables \ref{tab:supervised_ner}, \ref{tab:supervised_re}, \ref{tab:supervised_et}, and \ref{tab:supervised_eea}. We conclude two main findings: (1) SFT-basd methods provide LLMs with a comprehensive understanding of IE through training in large-scale IE datasets, achieving superior results on held-in tasks compared to RUIE, which utilizes a frozen LLM. However, RUIE achieved an F1 score of 92.26 on FindVehicle, showing the strong potential of retrieval-based UIE. (2) RUIE showed markedly superior performance to generic retrievers, with improvements of 6.47, 7.13, 1.35, and 4.98 on the NER, RE, ED, and EAE tasks, respectively. This shows the effectiveness of introducing large model preference and keyword enhancement during the retriever training process.

We have also identified several key error patterns: 1. Domain knowledge gaps: Poor performance on medical datasets (e.g. bg2gm, SciERC) due to LLaMA3.1's limited medical domain training. 2. Label confusion: Difficulty distinguishing similar target labels (e.g. FabNER, kbp37). 3. Text formatting: Decreased precision with massive special symbols such as ``@'' and ``\#'' in HarveyNER. 4. Pronoun handling: Missed pronoun entities in ACE2005. Moreover, we observed that the information extracted by the LLM and the actual labels only differ by one or two words, yet this does not hinder human understanding of the extraction results. This suggests that LLM-based information extraction methods require a more nuanced evaluation metric.

\begin{table}[ht]
\centering
\setlength{\tabcolsep}{2pt}
\begin{tabular}{l|c|c|c}
\toprule
LLMs & NER & RE & ED / EAE    \\
\midrule
RUIE	&	68.24	&	\textbf{51.55}	&	\textbf{50.84} / \textbf{40.16}	\\
- keyword-enhanced	&	\textbf{68.64}	&	51.37	&	48.82 / 39.47	\\
- reward model	&	63.69	&	32.24	&	43.57 / 35.83	\\
- distill loss	&	62.78	&	33.73	&	39.77 / 31.58	\\
\bottomrule
\end{tabular}
\caption{Ablation study across all held-out datasets. “- keyword-enhanced” trains the reward model and retriever without keyword enhancement. “- reward model” trains the retriever with distill loss on LLMs scores and contrastive loss on ranked candidates. “- distill loss” trains the retriever soly with contrastive loss on ranked candidates.}
\label{tab:ablation}
\end{table}

\subsection{Ablation Study}

We conducted an ablation study across all held-out datasets to underscore the effectiveness of the key innovations in our work (Table \ref{tab:ablation}). Removing the keyword enhancement led to a 0.62 decrease in the average F1-score across four tasks, highlighting the value of fine-grained information alignment during retriever training. After removing distill loss, the average F1-score of the four tasks decreases by 10.73, which is because the rank of candidate examples only reflects the relative distribution of LLMs preferences. However, knowledge distillation based on KL-divergence can align the absolute distribution of LLMs preferences, which confirms the importance of using LLMs preferences as supervision signals. After removing the reward model, the average F1-score of the four tasks decreases by 8.87, indicating the log-likelihood from LLMs is not suitable as a direct knowledge source for distillation. The reason is that the average log-likelihood is not a true probability distribution, and its values tend to cluster within a narrow range, making it less effective as a target distribution within the KL-divergence framework.

\subsection{The Effects of Different K-shot Demonstrations}

\begin{figure}[ht]
    \centering
    \includegraphics[width=0.8\linewidth]{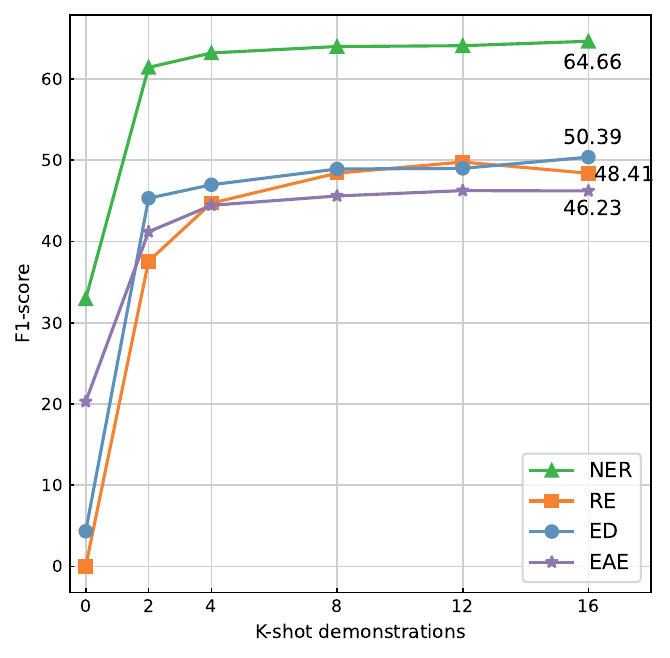}
    \caption{Performance (in F1-score) comparison by varying k-shot demonstrations.}
    \label{fig:k-shots}
\end{figure}

We first used the default experimental setup to investigate the effects of varying k-shot demonstrations on extraction performance. Experiments were conducted across all held-in and held-out datasets, and we reported the average F1 scores for each task across all datasets, as depicted in Figure \ref{fig:k-shots}. We summarized three findings: (1) The performance across all tasks improved with an increase in k, with the most notable enhancement occurring as k rose from 0 to 2, indicating that few-shot demonstrations are crucial for LLMs to tackle tasks effectively. (2) The task performance does not consistently improve with increasing k. For example, there is a performance drop when k increased from 12 to 16 in the RE task. We surmise that one contributing factor is the limitation of the model's context length, while another is the additional noise introduced by an excess of examples, which can adversely affect model performance. Therefore, we set k to 8, which can take into account both inference efficiency and extraction performance. (3) Model performance is influenced by the complexity of the task. NER has more data and is easier compared to other information extraction tasks, exhibited significantly better zero-shot performance than other tasks, while LLM was virtually incapable of completing the RE and ED tasks in a zero-shot scenario. Even as the number of examples k increased, the model's performance on NER remained superior to that of other information extraction tasks.

\begin{figure}[ht]
    \centering
    \includegraphics[width=0.8\linewidth]{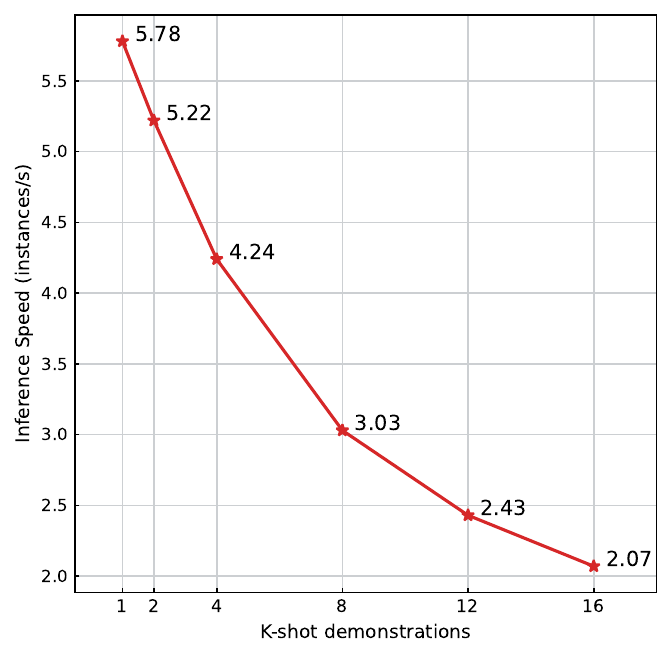}
    \caption{Inference speed (instances per second) comparison of different k-shots demonstrations on ACE 2004 (NER), ADE corpus (RE) and ACE2005 (ED and EAE) datasets.}
    \label{fig:k-shots_on_inference}
\end{figure}



Then we used the default experimental setup to investigate the effects of different k-shots on inference efficiency. Since SFT-based methods with similar size do not support vLLM, we treat the case of k-shot = 1 as equivalent to the SFT-based methods for fairness (RUIE is built based on bi-encoder and the retrieval time of a single sample can be negligible in our experiment). As shown in Figure \ref{fig:k-shots_on_inference}, the inference speed decreases consistently with increasing number of k shots, with k-shot = 1 approximately 1.9 times faster than k-shot = 8. While RUIE has lower inference efficiency than SFT-based due to retrieval overhead and longer context length, it offers three key advantages: 1. Minimal training cost: Only fine-tunes a small retriever vs. full LLM. 2. Better domain adaptation: Using ICL with a small amount of labeled data vs. full retraining. 3. API compatibility: Works directly with LLM APIs vs. requiring model deployment.

\subsection{The Effects of Different Scoring LLMs}

\begin{table}[h]
\centering
\setlength{\tabcolsep}{4pt}
\begin{tabular}{l|c|c|c}
\toprule
LLMs & NER & RE & ED / EAE    \\
\midrule
GPTNeo	&	\textbf{68.27}	&	50.78	&	47.86 / 40.35	\\
LLaMA3-instruct	&	68.14	&	51.24	&	49.07 / \textbf{40.68}	\\
LLaMA3	&	68.24	&	\textbf{51.55}	&	\textbf{50.84} / 40.16	\\
\bottomrule
\end{tabular}
\caption{Performance (in F1-score) comparison of different scoring LLMs on NER, RE, ED, EAE tasks across all held-out datasets.}
\label{tab:different_llms_scoring}
\end{table}

To investigate the impact of various scoring LLMs on the extraction performance, we conducted experiments across all held-out datasets, with results presented in Table \ref{tab:different_llms_scoring}. We have two findings: 1) The size of the scoring LLM has a minor influence on the final performance. Although LLaMA3 (8B) outperformed GPTNeo (2.7B) by 2.98 on ED task, their performance was comparable on NER, RE and EAE tasks. This suggests that models of different sizes exhibit similar preferences for certain tasks, allowing one to select an appropriately sized LLM based on computational resources. 2) The base version of the LLM is more suitable as a scoring model than the instruct version. We calculated the mean and variance of the scores for the positive samples (top-3) and the negative samples (last-16) for both versions of the LLM (Table \ref{tab:mean_and_variance}). Although the mean difference between positive and negative sample scores of instruct version LLM is larger, the variance of positive and negative sample scores is significantly larger than that of base version LLM, which introduces instability to the subsequent training of the reward model, resulting in the final extraction performance inferior to base version.

\subsection{Performance Analysis Across Different LLMs for Inference}

Table \ref{tab:different_llms_inference} shows the performance of various reasoning LLMs across all held-out datasets. The ability of the inference LLMs significantly influences the extraction performance, which is expressed in two aspects: 1) Model size: For the same Qwen1.5, an increase in model size from 7B to 14B resulted in an average performance enhancement of 9.21 across four tasks. 2) Model type: Within the LLaMA series, LLaMA3.1, which is an advancement over LLaMA3, achieved an average performance improvement of 0.87 across four tasks. Since all tasks are in English, the LLaMA series models consistently outperformed the Qwen series. Additionally, RUIE adapts effectively to both locally deployed models and API-based models, with Deepseek demonstrating the best performance overall.

\begin{table}[ht]
\centering
\begin{tabular}{l|c|c|c}
\toprule
LLMs & NER & RE & ED / EAE    \\
\midrule
Qwen1.5-7b	&	54.70	&	33.86	&	33.84 / 26.90	\\
Qwen1.5-14b	&	63.35	&	44.53	&	40.09 / 32.23	\\
LLaMA3-8b	&	67.19	&	53.65	&	46.83 / 39.64	\\
LLaMA3.1-8b	&	68.24	&	51.55	&	50.84 / 40.16	\\
Deepseek	&	\textbf{72.24}	&	\textbf{58.85}	&	\textbf{62.77} / \textbf{47.95}	\\
\bottomrule
\end{tabular}
\caption{Performance (in F1-score) comparison of different inference LLMs on NER, RE, ED, EAE tasks across all held-out datasets. Deepseek refers to the deepseek-chat-v2. All other models are instruct variant, which are further fine-tuned on instruction-tuning datasets after pre-training.}
\label{tab:different_llms_inference}
\end{table}

\section{Conclusion}
In this paper, we introduced RUIE, a novel trainable retrieval-based framework for unified information extraction that addresses key challenges in the field. Our framework introduced a pioneering trainable retrieval mechanism specifically designed for UIE tasks, significantly reducing computational costs while enabling rapid generalization to unseen tasks. Extensive experiments on 31 training datasets and 8 held-out tasks demonstrated RUIE's superior performance across various IE tasks, suggesting promising directions for developing more efficient and adaptable IE systems.

\section*{Limitations}
Sequence length constraint: RUIE currently focuses on sentence-level UIE. On the one hand, it is difficult to meet the retrieval requirements of long documents due to the length limitation of the retriever is 512. On the other hand, due to the length limitation of LLMs, k-shot cannot be increased in the case of long documents, which limits the performance of the model.

Gap between RUIE and the SFT-based methods in seen tasks: The SFT-based methods inject UIE-specific knowledge into the LLMs through fine-tuning, while RUIE frozens the LLMs and can only leverage the knowledge from the pre-training phase. How to better stimulate the information extraction ability of LLMs under controllable cost is a worthy study direction.

English corpus only: RUIE is currently only trained and tested on English data. In the future, we hope to expand RUIE to more languages.

\section*{Acknowledgments}
This work was supported in part by the National Key Research and Development Program of China (No.2021YFF1201200), the Science and Technology Major Project of Changsha (No.kh2402004). This work was carried out in part using computing resources at the High-Performance Computing Center of Central South University.

\bibliography{coling_main}

\appendix

\section{Data Details}
\label{sec:data_details}

Our candidate pool $P$ mainly consists of IE INSTRUCTIONS \cite{wang_instructuie_2023} and IEPILE \cite{gui-etal-2024-iepile}. There are 22 NER datasets: 
ACE2004 \cite{mitchell2005ace}, 
ACE2005 \cite{ace2005}, 
Broad Twitter~\cite{broad_twitter_corpus_DATASET},
CoNLL2003~\cite{CoNLL03_Dataset},
MultiNERD~\cite{multiNERD_DATASET}, 
Ontonotes~\cite{OntoNotesDataset},
Polyglot-NER~\cite{polyglot},
tweetNER7~\cite{tweetNER7DATASET},
wikiANN~\cite{wikiann_Dataset},
wikineural~\cite{wikineuralDATASET},
AnatEM~\cite{AnatEM_DATASET}, 
bc2gm~\cite{Kocaman2020BiomedicalNE_DATASET},
bc4chemd~\cite{bc4chemdDATASET},
bc5cdr~\cite{Li2016BioCreativeVC_DATASET}, 
FabNER\cite{Kumar2021FabNERIE}, 
FindVehicle~\cite{FindVehicle_DATASET},
GENIA~\cite{GENIANER_DATASET},
HarveyNER~\cite{HarveyNERDATASET},
MIT Movie~\cite{MITReviewDataset}
MIT Restaurant~\cite{MITReviewDataset}
ncbi-disease~\cite{ncbi-disease_DATASET}. For the RE task, we utilize 10 datasets including 
ADE corpus~\cite{ADEcorpus_DATASET}, 
CoNLL04~\cite{Roth2004ALP},
GIDS~\cite{Jat2018ImprovingDS},
kbp37~\cite{kbp37_DATASET}, 
NYT~\cite{NYT_DATASET},
NYT11 HRL~\cite{nyt11},
SciERC~\cite{SciERC_DATASET}, 
semeval RE~\cite{Hendrickx2010SemEval2010T8},
FewRel~\cite{FewRel_DATASET} and Wiki-ZSL~\cite{Wiki-ZSL_DATASET}. For the EE task, ACE2005~\cite{ace2005},  CASIE\cite{Lu2021Text2EventCS},
GENIA\cite{Kim2003GENIAC},
PHEE\cite{Sun2022PHEEAD},
CrudeOilNews~\cite{crudeoilnews},
RAMS~\cite{rams} and WikiEvents~\cite{wikievents} are used.

\section{Implementation Details}
\label{sec:implementation_details}
\begin{table}[ht]
\centering
\resizebox{0.85\linewidth}{!}{
\begin{tabular}{l|c|c}
\toprule
~ & Reward & Retriever \\
\midrule
learning rate	&	1e-5	&	3e-5	\\
batch size	&	64	&	128	\\
training steps	&	3000	&	6000	\\
$\alpha$	&	-	&	0.2	\\
$\tau$	&	-	&	0.01	\\
positives	&	top-3	&	top-3	\\
negatives	&	last-16	&	in-batch negatives	\\
input length	&	512	&	512	\\
\bottomrule
\end{tabular}
}
\caption{Hyperparameters for reward and retriever model training.}
\label{tab:hyperparameters}
\end{table}

We finished LLM scoring, reward and retriever training on two 3090 GPUs. In order to balance the efficiency and performance of the scoring process, we used bm25 to initialize 100 candidates for each sample in the training set. Due to the sequence length limitation of the LLMs, we retrieved 8 samples for each query and set the maximum input length to 1792 and the maximum generation length to 256. We performed inference using vLLM~\cite{vllm} on a single 3090 GPU. In order to ensure the reproduction of the results, we use a greedy decoding strategy and set the temperature to 0.

\section{Score Mean and Variance of different LLMs}

\begin{table}[hb]
\centering
\resizebox{0.85\linewidth}{!}{
\begin{tabular}{l|c|c|c|c}
\toprule
\multirow{2}*{Model} & \multicolumn{2}{c}{Positives} & \multicolumn{2}{c}{Negatives} \\
\cmidrule{2-5}
~ & mean & variance &  mean & variance \\
\midrule
base	&	-0.39	&  0.18	&	-1.02	&  0.93	\\
instruct	&	-0.36	&  0.35	&	-1.23	&  2.36	\\
\bottomrule
\end{tabular}
}
\caption{Score Mean and Variance of different LLMs. ``base'' indicates LLaMA3-8B and ``instruct'' indicates LLaMA3-8B-instruct. We reported score mean and variance in positives and negatives separately.}
\label{tab:mean_and_variance}
\end{table}


\begin{table*}[ht]
\centering
\begin{tabular}{c|c|c|c|c|c|c}
\toprule
Task & Dataset & \#Schema & \#Train & \#Test & Training & Evaluation    \\
\midrule
\multirow{26}*{NER}	&	ACE2004	&	7	&	6202	&	812	&	$\surd$	&	$\surd$	\\
~	&	ACE2005	&	7	&	7299	&	1060	&	$\surd$	&	$\surd$	\\
~	&	Broad Tweet	&	3	&	5334	&	2001	&	$\surd$	&	$\surd$	\\
~	&	CoNLL2003	&	4	&	14041	&	3453	&	$\surd$	&	$\surd$	\\
~	&	multiNERD	&	16	&	134144	&	10000	&	$\surd$	&	$\surd$	\\
~	&	Ontonotes	&	18	&	59924	&	8262	&	$\surd$	&	$\surd$	\\
~	&	Polyglot-NER	&	3	&	393982	&	10000	&	$\surd$	&	$\surd$	\\
~	&	tweetNER7	&	7	&	7111	&	576	&	$\surd$	&	$\surd$	\\
~	&	Wikiann	&	3	&	20000	&	10000	&	$\surd$	&	$\surd$	\\
~	&	wikineural	&	3	&	92729	&	11597	&	$\surd$	&	$\surd$	\\
~	&	anatEM	&	1	&	5861	&	3830	&	$\surd$	&	$\surd$	\\
~	&	Bc2gm	&	1	&	12500	&	5000	&	$\surd$	&	$\surd$	\\
~	&	Bc4chemd	&	1	&	30682	&	26364	&	$\surd$	&	$\surd$	\\
~	&	Bc5cd	&	2	&	4560	&	4797	&	$\surd$	&	$\surd$	\\
~	&	FabNER	&	12	&	9435	&	2064	&	$\surd$	&	$\surd$	\\
~	&	FindVehicle	&	21	&	21565	&	20777	&	$\surd$	&	$\surd$	\\
~	&	GENIA	&	5	&	15023	&	1854	&	$\surd$	&	$\surd$	\\
~	&	HarveyNER	&	4	&	3967	&	1303	&	$\surd$	&	$\surd$	\\
~	&	Ncbi-disease	&	1	&	5432	&	940	&	$\surd$	&	$\surd$	\\
~	&	CrossNER AI$\dagger$	&	14	&	-	&	431	&		&	$\surd$	\\
~	&	CrossNER Literature$\dagger$	&	12	&	-	&	416	&		&	$\surd$	\\
~	&	CrossNER Music$\dagger$	&	13	&	-	&	465	&		&	$\surd$	\\
~	&	CrossNER Politics$\dagger$	&	9	&	-	&	650	&		&	$\surd$	\\
~	&	CrossNER Science$\dagger$	&	17	&	-	&	543	&		&	$\surd$	\\
~	&	MIT Movie Review$\dagger$	&	12	&	-	&	2442	&		&	$\surd$	\\
~	&	MIT Restaurant Review$\dagger$	&	8	&	-	&	1520	&		&	$\surd$	\\
\midrule
\multirow{10}*{RE}	&	ADE corpus	&	1	&	3417	&	428	&	$\surd$	&	$\surd$	\\
~	&	CoNLL2004	&	5	&	922	&	288	&	$\surd$	&	$\surd$	\\
~	&	GIDS	&	4	&	8526	&	4307	&	$\surd$	&	$\surd$	\\
~	&	Kbp37	&	18	&	15917	&	3405	&	$\surd$	&	$\surd$	\\
~	&	NYT	&	24	&	56196	&	5000	&	$\surd$	&	$\surd$	\\
~	&	NYT11 HRL	&	12	&	62648	&	369	&	$\surd$	&	$\surd$	\\
~	&	SciERC	&	7	&	1366	&	397	&	$\surd$	&	$\surd$	\\
~	&	Semeval RE	&	10	&	6507	&	2717	&	$\surd$	&	$\surd$	\\
~	&	FewRel$\dagger$	&	83	&	-	&	17291	&		&	$\surd$	\\
~	&	Wiki$\dagger$	&	100	&	-	&	23113	&		&	$\surd$	\\
\midrule
\multirow{7}*{EE}	&	ACE2005	&	33(22)	&	3342	&	293	&	$\surd$	&	$\surd$	\\
~	&	CASIE	&	5(26)	&	3751	&	1500	&	$\surd$	&	$\surd$	\\
~	&	GENIA	&	5(0)	&	15023	&	1854	&	$\surd$	&	$\surd$	\\
~	&	PHEE	&	2(16)	&	2898	&	968	&	$\surd$	&	$\surd$	\\
~	&	CrudeOilNews$\dagger$	&	18(104)	&	-	&	356	&		&	$\surd$	\\
~	&	RAMS$\dagger$	&	106(398)	&	-	&	887	&		&	$\surd$	\\
~	&	WikiEvents$\dagger$	&	31(81)	&	-	&	249	&		&	$\surd$	\\
\bottomrule
\end{tabular}
\caption{Statistical data of all 39 IE datasets. Datasets with $\dagger$ are held-out datasets, which is unseen during retriever training stage. CrossNER~\cite{CrossNERDATASET} is divided into five subsets for our statistical analysis. For the held-out datasets, we only report their test set.}
\label{tab:dataset_details}
\end{table*}

\section{Sample Format}
\label{sec:sample_format}

Detailed sample format is listed in Table~\ref{tab:sample_format}.

\section{Instruction Format}
\label{sec:instruction_format}

Detailed instruction format is listed in Table~\ref{tab:instruction_format}.

\begin{table*}[ht]
\centering
\begin{tabular}{c|p{0.9\textwidth}}
\toprule
Task    &   \multicolumn{1}{c}{Sample Format} \\
\midrule
\multirow{4}*[-0.1in]{NER} &    \textbf{Task:} Named Entity Recognition   \\
~ & \textbf{Schema:} [location, person, organization] \\
~ & \textbf{Input:} The Parkinsons are a punk rock band originally from Coimbra, Portugal, formed in the year 2000 and based in London, known for their outrageous live performances.    \\
~ & \textbf{Output:} location: Coimbra; location: Portugal; location: London.    \\
\midrule
\multirow{4}*[-0.1in]{RE} &    \textbf{Task:} Relation Extraction   \\
~ & \textbf{Schema:} [Organization based in, Located in, Live in, Work for, Kill] \\
~ & \textbf{Input:} Washington: About 110 firefighters cut a containment line most of the way around an 850-acre forest fire in the Pasayten Wilderness near the Canadian border Tuesday.    \\
~ & \textbf{Output:} Located in: Pasayten Wilderness, Washington.    \\
\midrule
\multirow{4}*[-0.2in]{ED} &    \textbf{Task:} Event Detection   \\
~ & \textbf{Schema:} [phishing, data breach, ransom, discover vulnerability, patch vulnerability] \\
~ & \textbf{Input:} Google Project Zero's security researchers have discovered another critical remote code execution (RCE) vulnerability in Microsoft’s Windows operating system, claiming that it is something truly bad.    \\
~ & \textbf{Output:} discover vulnerability: have discovered.    \\
\midrule
\multirow{4}*[-0.5in]{EAE} &    \textbf{Task:} Event argument extraction   \\
~ & \textbf{Schema:} Given event trigger: ``discover vulnerability: discovered''; Candidate arguments: [vulnerable system owner, vulnerability, capabilities, time, vulnerable system version, discoverer, common vulnerabilities and exposures, supported platform, vulnerable system] \\
~ & \textbf{Input:} Google Project Zero's security researchers have discovered another critical remote code execution (RCE) vulnerability in Microsoft’s Windows operating system, claiming that it is something truly bad.    \\
~ & \textbf{Output:} vulnerable system: Windows operating system; vulnerability: remote code execution (RCE) vulnerability; discoverer: security researchers; vulnerable system owner: Microsoft; discoverer: Google Project Zero.    \\
\bottomrule
\end{tabular}
\caption{Detailed sample format across four IE tasks. NER sample is from polyglotNER~\cite{polyglot}, RE sample is from conll04~\cite{Roth2004ALP}, and both ED and EAE samples are from CASIE~\cite{Lu2021Text2EventCS}.}
\label{tab:sample_format}
\end{table*}

\begin{table*}[ht]
\centering
\begin{tabular}{c|p{0.9\textwidth}}
\toprule
Task    &   \multicolumn{1}{c}{Instruction} \\
\midrule
\multirow{5}*[-0.35in]{NER} &    Please analyze the given schema and extract all named entities from the provided input. Follow these instructions carefully:   \\
~ & 1. Output Format: Present the extracted entities in this structured format: ``EntityType1: EntityName1; EntityType2: EntityName2; ...''. \\
~ & 2. Include Only Present Entities: Only output entities that actually exist in the input. Ignore any entities that are not mentioned.    \\
~ & 3. No Entities Response: If the input contains no named entities, respond with ``None''.    \\
~ & 4. Examples Handling: If examples are provided for reference, use them to understand the annotation criteria, but do not extract entities from these examples.  \\
\midrule
\multirow{5}*[-0.35in]{RE} &    Please analyze the given sentence and extract subjects and objects that have a specific relation, according to the provided schema. Follow these guidelines:   \\
~ & 1. Output Format: Format your output as follows: ``relation1: subject1, object1; relation2: subject2, object2; ...''. \\
~ & 2. Include Only Present Relations: Only output relations that actually exist in the input. Ignore any relations that are not mentioned.    \\
~ & 3. No Relations Response: If the sentence contains no relations, respond with ``None''.    \\
~ & 4. Example Usage: If examples are provided, use them to understand the annotation criteria. Do not extract relations from these examples.  \\    
\midrule
\multirow{5}*[-0.3in]{ED} &    Please analyze the given schema and extract all event triggers from the provided input. Follow these instructions carefully:   \\
~ & 1. Output Format: Present the extracted triggers in this structured format: ``EventType1: TriggerName1; EventType2: TriggerName2; ...''. \\
~ & 2. Include Only Present Triggers: Only output triggers that actually exist in the input. Ignore any triggers that are not mentioned.    \\
~ & 3. No Triggers Response: If the input contains no event triggers, respond with ``None''.    \\
~ & 4. Examples Handling: If examples are provided for reference, use them to understand the annotation criteria, but do not extract triggers from these examples.  \\
\midrule
\multirow{5}*[-0.45in]{EAE} &    Please extract and list arguments of specified types for a given event type and trigger, according to the provided schema. Follow these instructions carefully:   \\
~ & 1. Output Format: Format your output as follows: ``ArgumentType1: argument1; ArgumentType2: argument2; ...''. \\
~ & 2. Include Only Present Arguments: Only include event arguments that are present in the input sentence. Ignore any event arguments that are not mentioned.    \\
~ & 3. No Arguments Response: If the sentence contains no event arguments, respond with ``None''.    \\
~ & 4. Example Usage: If examples are provided, use them to understand the annotation criteria. Do not extract event arguments from these examples.  \\
\bottomrule
\end{tabular}
\caption{Detailed instruction format across four IE tasks.}
\label{tab:instruction_format}
\end{table*}

\end{document}